# BOOSTING-LIKE DEEP LEARNING FOR PEDESTRIAN DETECTION

Lei Wang[1,1], Baochang Zhang [1]

[1]School of Automation Science and Electrical Engineering

Beihang University, Beijing, China

wtiffanyl@163.com; bczhang@buaa.edu.cn

**Abstract.** This paper proposes boosting-like deep learning (BDL) framework for pedestrian detection. Due to overtraining on the limited training samples, overfitting is a major problem of deep learning. We incorporate a boosting-like technique into deep learning to weigh the training samples, and thus prevent overtraining in the iterative process. We theoretically give the details of derivation of our algorithm, and report the experimental results on open data sets showing that BDL achieves a better stable performance than the state-of-the-arts. Our approach achieves 15.85% and 3.81% reduction in the average miss rate compared with ACF and JointDeep on the largest Caltech benchmark dataset, respectively.

## 1  Introduction

Pedestrian detection has an important significance in real life, which has been widely used in intelligent control systems, traffic safety assist systems, robotics research and other fields. It has attracted more and more attention, and a variety of feature extraction and classification methods have been proposed. The main feature extracting methods are divided into two categories: handcrafted extraction and automatic learning. The famous handcrafted extraction methods are HOG [1]. HOG portrays the local gradient magnitude and direction of the image, which normalizes vector feature blocks based on gradient features. It allows overlap between blocks, thus it is not sensitive to the light changes and a small amount shift. Therefore, it can effectively depict the human body edge feature. Other commonly used methods are Haar-like [2], SIFT [3], covariance descriptors [4], integral channel features [5], 3D geometric characteristic [6] and so on. With the computer development and data volumes grow, automatic learning methods are gradually put forward. Sermanet et al. [7] proposed the ConvNet structure, which uses the original pixel values as the input. It combines unsupervised and supervised methods to train multi-stage automatic sparse convolution encoder. It shows a relatively impressive result on INRIA pedestrian database, but does not give test results on Caltech. UDN [8] is a joint deep neural network model combined with deformation and occlusion model, which achieves a lower miss rate than the conventional human detection HOGCSS SVM algorithm on Caltech and ETH dataset. Lim J et al. [9] use a supervised training mode to extract middle class feature based on contour information, and train random forest classifier to improve performance.

The manual feature extraction has a good description for pedestrians, but it can't learn the essential characteristics and has poor adaptability. The latter can automatically extract pedestrian features by methods such as feedback propagation. But it requires a lot of training samples and takes a lot of time. What's more, it has higher hardware requirement. Based on the characteristics of the two kinds of feature extraction methods, we propose a novel pedestrian detection framework. It combines manual feature extraction method with the deep learning model, and incorporates the boost ideas into our framework. We gradually adjust the sample weight in feedback training process. Our method can not only improve pedestrian detection performance, but also strengthen the stability of the convolution neural network. In addition, our input features are inspired by integral channel features [10], but we regularize them in order to improve the detection rate. Our deep structure only uses a simple convolutional neural network [11] which consists of two convolutional layers to gain a higher level feature expression. The final classifier we use is only a simple single neural network.

The main contributions of this paper are described as follows: firstly, the combination of modified integral channel features and simple CNN network, using the back-propagation algorithm for training; secondly, the introduction of boosting-like structure in the deep learning network, gradually adjusting the training sample weights in the feedback propagation to improve the detection structure stability and convergence speed.

The remainder of this paper is arranged as follows: in section 2, our pedestrian detection structure is described, including the input channel features and deep learning structure. In section 3, the boosting-like algorithm that we proposed is elaborated. In section 4, experimental results are presented and analyzed, and in section 5, we conclude the paper.

## 2    Pedestrian Detection Framework

### 2.1    Input channel features

Channel features can be gained through making different output functions in response to input image I which may be linear or non-linear transformation, such as the Gabor filters belonging to the linear transformation, and canny edge detection non-linear transformation. Assume channel response function is f, the channel output is C, the image I respond is given as follows:

$$C = f(I) . \qquad (1)$$

where f denotes a simple first-order feature function as a sum of pixels in a fixed rectangular area. High-order features can be obtained by combining several first-order functions via a variety of strategies. In addition, the combination of different channels is denoted the aggregate channel features. In this paper, we extract low-level pedestrian features in order to improve the training speed by reducing the number layers of deep learning network, and then extract the high-level features through our deep network. Specific steps are as follows:

Step one: In this paper, we use color and gradient feature as input channels. In concrete terms, we change the RGB input image into LUV three-channel image, then one gradient magnitude channel (| G |) and six histogram of gradient oriented channels(G1-G6) are gained through conversion processing on input images.

Step two: Since pedestrian images are influenced seriously by illumination, the data in each channel is processed to be zero mean and unit variance. Since our network activation function is Sigmoid, this processing can also increase the convergence rate in the gradient descent process [12].

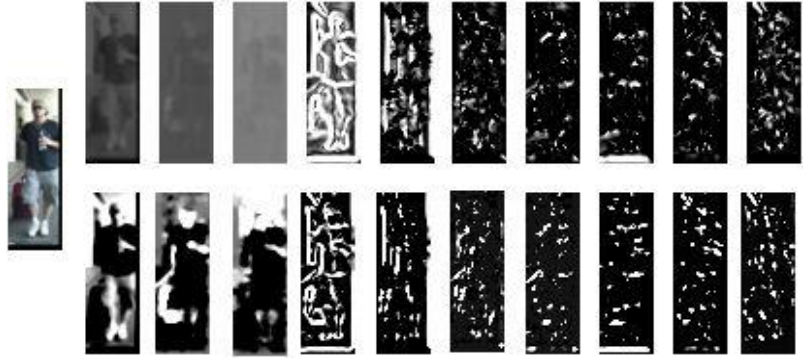

**Fig. 1.** Comparison of channel features

The first column is the original input image, then the second column to the tenth column correspond to LUV, | G | and the G1-G6 channel features. The first row is the original ACF feature, and the second row is the one after normalization. The results show that regularization not only enhances the image resolution significantly, but also highlights the pedestrian details.

**2.2 CNN structure**

CNN is the neural network including multilayer, and each layer includes a plurality of two dimensional planes. Each plane also includes multiple neurons. The two-dimensional image data can be directly as input channels. Furthermore, feature extraction step has been embedded in the structure of CNN.

CNN mainly achieves deformation, shift and scale invariance [13] by local receptive fields, shared weights and down-sampling. The local receptive fields of two dimensional -space can make neural network extract primary visual features from the input image, such as edge, endpoint, corner and so on. Subsequent layers obtain higher level features through combining these primary characteristics. Neurons can detect the same characteristics in different locations on the input image by shared weights. Then the input translation changes will be appeared in the output in the same direction and distance, but it doesn't cause other forms of change. At the same time, the weight sharing also significantly reduces the training weights number. Sub-sampling not only filters out the noisy characteristics, but also enhances features which are crucial to image recognition. CNN model in this paper is shown in Figure 2:

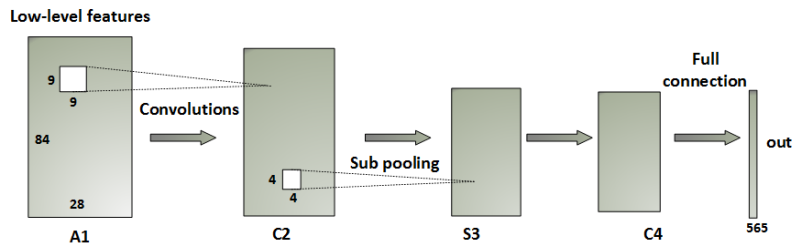

**Fig. 2.** CNN structure

As Figure 2 shown, there are total five layers in addition to input layer. The input image size is 84 x 28 pixels. A1 layer consists of 10 feature maps, which size is the same as the input one 84 x 28. C2 layer contains 64 filter kernels, and each kernel includes 9 x 9 = 81 weight parameters. It's multiplied by 10 and adding a bias, so there are (9x9 x 10 +1 ) x 64 = 61904 parameters. The 10 input channels respectively convolution with 64 filters, then add a bias. All of them are entered into an excitation function. It is calculated by the following formula:

$$M_k^l = S\left(\sum_{n \in I_k} \omega_{nk} * M_n^{l-1} + b_n^l\right). \tag{2}$$

Where $M_k^l$ denotes the k-th feature map of the l-th layer, and $I_k$ represents all the feature maps of the k-th channel. $\omega_{nk}$ indicates learning parameter that corresponds to convolutional kernel, and $b_n^l$ represents the bias of the n-th input image in the l-th layer. S (·) is the activation function, such as sigmoid function. S3 is the down-sampling layer, wherein each neuron in the feature map corresponds to 4 x 4 neighborhood of C2 layer. 16 units are summed in S3 layer, and multiplied by a training parameter with a training bias. Finally, they are transferred by an excitation function to obtain 64 feature maps with size 21x7. It is calculated as follows:

$$M_k^l = S\left(\beta \sum_{n \in I_k} M_n^{l-1} + b_n^l\right). \tag{3}$$

Here β represents scalar training parameters which values vary with the sub-sampling methods, such as using Mean-Pooling, β = 1 / m, m represents down sample in m × m pixels (common size of 2 × 2). So the output image size is reduced to m times of original image. The output map has a bias denoted by $b_n^l$, then it is transferred into a nonlinear function (such as Sigmoid function).

C4 is also a convolutional layer, and we use 20 filter kernels with different size, and gain 20 different size feature maps. When cascading all of them, we obtain a final fully connected layer. The number is 565 in the full connection layer of our network. Finally, it is the recognition classifier, which should be differentiable on weights. Only in this case, you can use BP algorithm to train the network. CNN classifier used in this paper is a single fully connected neural network, other commonly used function such as logistic regression polynomial or radial basis function.

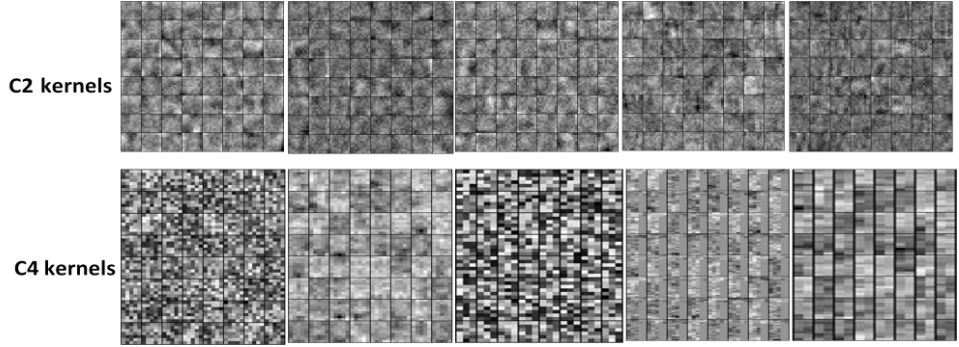

**Fig. 3.** Visualization of convolution kernel maps

The first row is the visual maps of the second layer convolution kernels, and the second row is the visual maps of the fourth layer convolution kernels. It can be seen that the filter kernels of the second layer mostly contain edges, lines features, and the fourth layer mainly consists of corner, point features. Deep network extracts more and more essential features with layers increasing, which can describe the pedestrian nature characteristics.

## 3   Boosting-like

As we all know, the training method of the convolutional neural network is mostly back propagation, and the network stability and convergence speed is a common problem in the process of training the model. If the learning rate is too large to gain a fast convergence speed it's easy to fall into local minimum. Otherwise, slow updating could result in time-consuming [14]. Therefore, we consider the boost algorithm, which adjusts the update rate according to the samples classification situation in the training process. It ensures the convergence speed and prevents network overfitting, which makes the network more stable. Generally, the cost function of convolution neural network is square error function. Assuming the samples are divided into C classes, the individual error for n-th samples is given by:

$$E = \frac{1}{2}\sum_{k=1}^{c}(t_k^n - y_k^n)^2 . \tag{4}$$

Here $t_k^n$ denotes the k-th dimension of the n-th sample label, and $y_k^n$ is similarly the k-th output layer unit in response to the n-th sample. In the practical application, we always sum the square error of all samples. The input $u^\ell$ of the $\ell$th layer and output $x^{\ell-1}$ of $(\ell-1)$th layer have the following linear relationship:

$$u^\ell = \alpha w^\ell x^{\ell-1} + b^\ell, \quad x^\ell = f(u^\ell). \tag{5}$$

Where $w^\ell$ denotes output layer weight and $b^\ell$ denotes the bias. They are constantly adjusted in the training process. $x^{\ell-1}$ is the $\ell-1$ layer output, and $\alpha$ is a penalty weight. f is the excitation function for the output layer, which is commonly chosen to be the sigmoid function or hyperbolic tangent function. Then the output layer sensitivity by the derivation is:

$$\delta^\ell = f'(u^\ell) * (y^n - t^n)\alpha . \tag{6}$$

The derivative of error E against weight $W^\ell$ is as follows:

$$\frac{\partial E}{\partial w^\ell} = x^{\ell-1}(\delta^\ell)^T = x^{\ell-1}f'(u^\ell) * (y^n - t^n)\alpha . \tag{7}$$

Finally, the delta updating rule is applied to each neuron to gain the new weights. The formula is given by:

$$w^{\ell+1} = w^\ell - \eta\frac{\partial E}{\partial w^\ell} = w^\ell - \eta x^{\ell-1}f'(u^\ell) * (y^n - t^n)\alpha . \tag{8}$$

Here $\eta$ is the learning rate, thus we can obtain the weight w updating method. Actually, the convolution neural network itself can be seen as several cascaded feature extractors, and each layer can be considered as a feature extractor. The features extracted are from low-level to high-level, and the results have a mutual suppression, which is to say that a classifier output not only has a relationship with the previous layer but also the next one. According to the formula (8), we distribute the feedback weights of right and wrong classification samples, which is feedback propagated from the last layer to the beginning layer.

$$Od_{t+1} = \begin{cases} (O_t - Y_t)\alpha_r, & Od_t < 0.5 \\ (O_t - Y_t)\alpha_w, & Od_t \geq 0.5 \end{cases} . \tag{9}$$

Where $Od_t$ is the output error, $O_t$ is the actual detection value of the network, and $Y_t$ is the sample target value. Meanwhile $Od_t$ is the output layer sensitivity $\delta$ in our network. $\alpha_r$ and $\alpha_w$ are respectively the penalty coefficients of right and wrong classified samples. When the sample output value is different with its label, the penalty weight should be increased; on the contrary, when the same, it should be decreased. This idea is similar to the boost algorithm, which trains different classifier by constantly updating the weights of training samples. It can avoid overfitting, thus making the network performance more stable. In this paper, the overall pedestrian detection framework is shown in Figure 4.

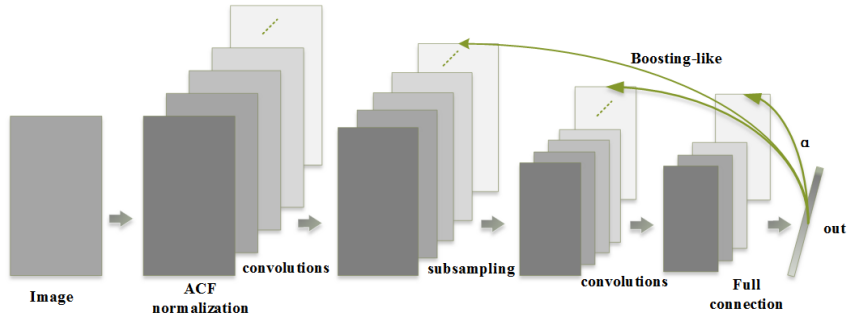

**Fig. 4.** Our pedestrian detection framework

Overview of our pedestrian detection structure, original image is extracted with ACF features and regularized. Then through the two layer convolution operations higher level pedestrian features are obtained. According to the output value of sample, the boost theory is merged into convolutional neural network over feedback propagation. This method not only enhances the network stability, but also greatly improves the detection performance.

## 4 Experimental results

Our pedestrian detection framework is evaluated on the Caltech dataset, which is currently the commonly used pedestrian dataset. It contains many complicated scenes including occlusion, illumination, deformation, and so on. In the experiment we use set00~set05 to train our model. There are about 60000 training samples, which include about 4000 positive ones. Set06~set10 are adopted as test sets. Usually sliding windows are used to traverse the pedestrian image in the detection stage. It is well known that the feedback process in the deep learning network structure takes a lot of time. Therefore, we use the strategy which is similar to UDN [8]. The detector using HOG+CSS and linear SVM is utilized for pruning candidate detection windows to save computation, and then the candidate windows are detected by deep network structure. These candidate windows have a high recall rate, and certainly contain a lot of false positive windows. This approach not only improves the training speed, but also meets the needs on testing the performance of our pedestrian detection structure.

### 4.1 Comparison of boosting-like stability

Figure 5 shows the comparison of using boosting-like algorithm or not in pedestrian detection structure. The abscissa represents training iterations, and the ordinate denotes the average miss rate which is tested by every trained model on Caltech-test dataset. This log-average miss rate is evaluated by the unified criteria proposed in [15]. Furthermore, the experiments are carried out on Caltech database.

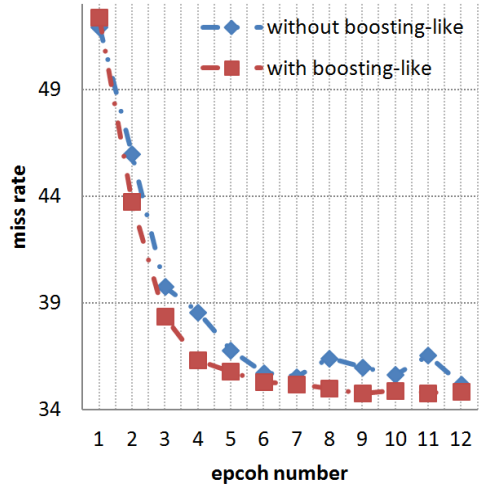

**Fig. 5.** Comparison of Boosting-like in terms of stability and detection performance

We can see from the comparison results: firstly, the curve is relatively stable when using boosting-like method in the feedback propagation, and volatility is smaller. While it's poor in stability without the boosting-like, volatility is large. Secondly, the boosting-like algorithm achieves 0.48% pedestrian detection performance gain. Therefore, this method not only improves the network stability, but also slightly improves the system detection accuracy while it doesn't reduce the convergence speed.

### 4.2 Results of the Caltech dataset

The evaluation method proposed in [15] is used to check detection performance of our pedestrian detection framework which gains the curve between the log-average miss rate and false positive rate of each image. In the experiment, we evaluate the detection performance in the reasonable subset which is a commonly used pedestrian detection collection. It consists of pedestrians who are more than 49 pixels in height, and whose occluded portions are less than 35%. We compared with the popular approaches related to our method: VJ [16], HOG, ConvNet [17], ACF [18], JointDeep [8], HOGCSS. These methods use various features, deformation models and different classifiers. Our experimental method is denoted by BDL.

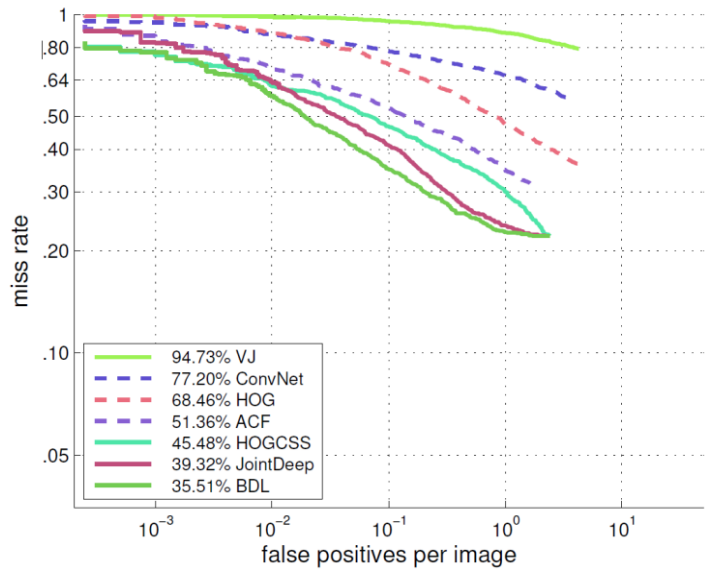

**Fig. 6.** Comparison of *log-average miss rate* vs. *false positives per image (FPPI)* between our approach *BDL* and related methods on Caltech dataset.

Figure 6 shows that the log-average miss rate of the pedestrian detection method we propose is 35.51%. It can be seen that our approach gets 15.85% and 3.81% performance gains compared with the ACF [10] and JointDeep [8] respectively on Caltech test. It should be noted that complicated methods such as deformation model, occlusion model, context information and joint training are not employed in our framework. Figure 7 shows some of pedestrian detection results on Caltech dataset.

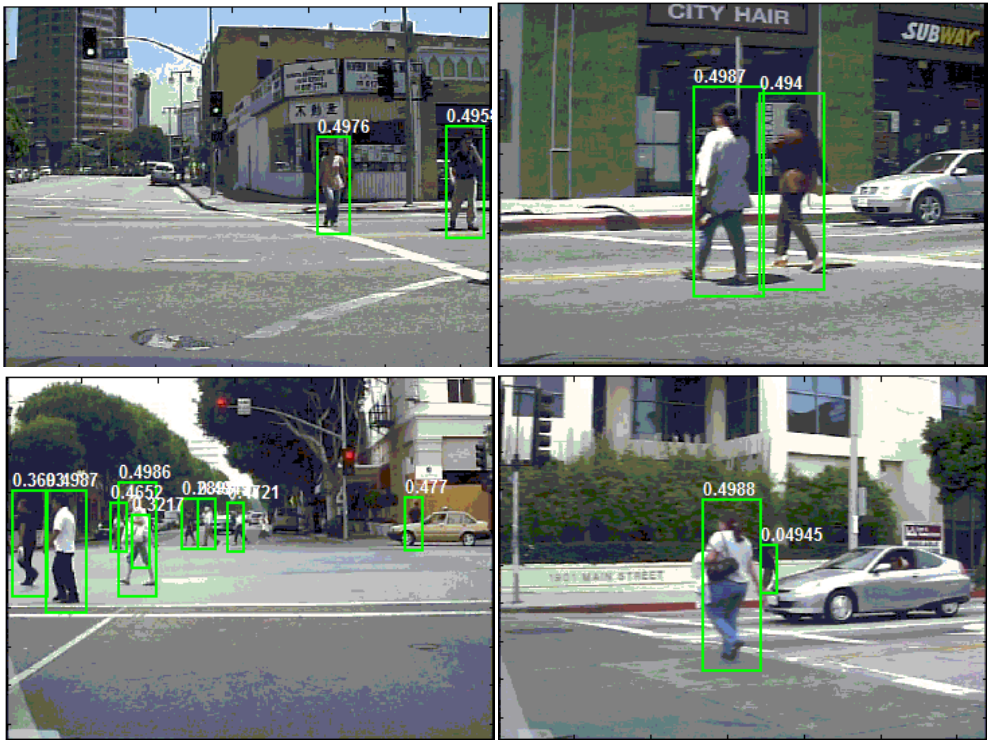

**Fig. 7.** Samples of pedestrian detection results on Caltech dataset

## 5   Conclusion

In this paper, we propose a simple but effective pedestrian detection framework. We use a similar boost idea to train the network structure which not only improves the system stability but also reduces the average miss rate of pedestrian detection. Through interaction, it achieves 15.85% and 3.81% performance gains compared with the corresponding methods on the largest Caltech dataset, respectively. Finally the experimental results demonstrate the validity and stability of our model. Through the experiments we can get the following conclusions: boosting-like method that we propose can improve detection stability and performance. The deep model is more time-consuming while training process, while the traditional handcrafted feature extraction owns poor adaptability. When we use low- level handcarfted features as input channels, less layers of deep structure can effectively improve the classification performance, meanwhile it can improve the training speed.  We are certain that if the pedestrian detection measures such as momentum [19], dropout [20] or multi-scale [21] techniques are carried out by our structure, the system performance will further enhance.

# References

1. Dalal N, Triggs B: Histograms of Oriented Gradients for Human Detection[C]. //IEEE Conference on Computer Vision & Pattern Recognition. IEEE Computer Society, 886--893 (2005)
2. P. Viola, M. J. Jones, and D. Snow: Detecting pedestrians using patterns of motion and appearance. IJCV, 63(2):153--161(2005)
3. A. Vedaldi, V. Gulshan, M. Varma, and A. Zisserman: Multiple kernels for object detection. IEEE 12th International Conference, 606--613(2009)
4. O Tuzel, F Porikli, P Meer: Pedestrian detection via classification on riemannian manifolds. IEEE Trans.PAMI, 30(10):1713--1727(2008)
5. P. Doll´ar, Z. Tu, P. Perona, and S. Belongie: Integral channel features. In: BMVC, 2009, vol. 2, p.(2009)
6. Hoiem D, Efros A A, Hebert M: Putting objects in perspective[J]. International Journal of Computer Vision, 80(1):2137--2144(2006)
7. Sermanet P, Soumithchintala K, Lecun Y: Pedestrian Detection with Unsupervised Multi-Stage Feature Learning[J]. IEEE Conference on Computer Vision & Pattern Recognition, 3626--3633(2012)
8. Wanli Ouyang,Xiaogang Wang: Joint Deep Learning for Pedestrian Detection[C]. ICCV,266--274(2013)
9. Lim J J, Lawrence Zitnick C, Dollár P. Sketch Tokens: A Learned Mid-level Representation for Contour and Object Detection[J]. IEEE Conference on Computer Vision & Pattern Recognition, 9(4):3158--3165(2013)
10. Dollar P, Appel R, Belongie S, et al: Fast Feature Pyramids for Object Detection[J]. IEEE Transactions on Pattern Analysis & Machine Intelligence, 36(8):1532--1545(2014)
11．LeCun Y,Bottou L, Bengio Y, et al: Gradient-based learning applied to document recognition[J]. Proc of the IEEE,86(11): 2278—2324(1998)
12. Bouvrie J, Bouvrie J: Notes on Convolutional Neural Networks[J]. Neural Nets(2006)
13. HUBEL D H, WUESEK T N. Receptive fields, binocular interaction and functional architecture in the cat's visual cortex[J]. J. Physiol, 1962, 160(12), 106--154(1962)
14. CHEN Y N, HAN C C, WANG C T, et al.:The application of a convolution neural network on face and license plate detection[C]. Proc. 18th Int. Conf. Pattern Recognition (ICPR'06), 552--555(2006)
15. P. Doll´ar, C. Wojek, B. Schiele, P. Perona: Pedestrian detection: An evaluation of the state of the art. IEEE Transactions on Software Engineering, 34(4):743--761(2012)
16. P. Viola, M. J. Jones, and D. Snow: Detecting pedestrians using patterns of motion and appearance. IJCV, 63(2):153--161(2005)
17. Sermanet P, Kavukcuoglu K, Chintala S, et al.: Pedestrian Detection with Unsupervised Multi-Stage Feature Learning[J]. IEEE Conference on Computer Vision & Pattern Recognition, 3626--3633(2012)
18. Nam W, Dollr P, Han J H: Local Decorrelation For Improved Detection[J]. Eprint Arxiv(2014)
19. Sutskever I, Martens J, Dahl G, et al: On the importance of initialization and momentum in deep learning[J]. Proceedings of International Conference on Machine Learning (2013)
20. Baldi P, Sadowski P: The Dropout Learning Algorithm.[J]. Artificial Intelligence, 210(3):78--122 (2014)
21. Ye Q, Jiao J, Zhang B: Fast pedestrian detection with multi-scale orientation features and two-stage classifiers[C]. IEEE International Conference on Image Processing, 881--884(2010)